\documentclass[sigconf]{acmart}

\usepackage{booktabs} 
\usepackage{breqn}
\usepackage{graphicx}
\usepackage{mathrsfs}
\usepackage{multirow}
\usepackage{amsmath}
\usepackage{setspace}

\setcopyright{acmcopyright}

\copyrightyear{2017}
\acmYear{2017}
\setcopyright{acmcopyright}
\acmConference{KDD '17}{August 13-17, 2017}{Halifax, NS, Canada}\acmPrice{15.00}\acmDOI{10.1145/3097983.3098011}
\acmISBN{978-1-4503-4887-4/17/08}

\begin{document}
\title{Cascade Ranking for Operational E-commerce Search}

\author{Shichen Liu, Fei Xiao}
\authornote{Both authors contributed equally to this study}
\affiliation{%
  \institution{Alibaba Group}
  \city{Hangzhou} 
  \state{China} 
}
\email{shichen.lsc@alibaba-inc.com}
\email{guren.xf@alibaba-inc.com}

\author{Wenwu Ou}
\affiliation{%
  \institution{Alibaba Group}
  \city{Hangzhou} 
  \state{China} 
}
\email{santong.oww@alibaba-inc.com}

\author{Luo Si}
\affiliation{%
  \institution{Alibaba Group}
\city{Seattle \& Hangzhou} 
\state{China} 
}
\email{luo.si@alibaba-inc.com}

\renewcommand{\shortauthors}{S. Liu et al.}

\begin{abstract}
		In the ``Big Data'' era, many real-world applications like search involve the ranking problem for a large number of items. It is important to obtain effective ranking results and at the same time obtain the results efficiently in a timely manner for providing good user experience and saving computational costs. Valuable prior research has been conducted for learning to efficiently rank like the cascade ranking (learning) model, which uses a sequence of ranking functions to progressively filter some items and rank the remaining items. However, most existing research of learning to efficiently rank in search is studied in a relatively small computing environments with simulated user queries.

This paper presents novel research and thorough study of designing and deploying a Cascade model in a Large-scale Operational E-commerce Search application (CLOES), which deals with hundreds of millions of user queries per day with hundreds of servers. The challenge of the real-world application provides new insights for research: 1). Real-world search applications often involve multiple factors of preferences or constraints with respect to user experience and computational costs such as search accuracy, search latency, size of search results and total CPU cost, while most existing search solutions only address one or two factors; 2). Effectiveness of e-commerce search involves multiple types of user behaviors such as click and purchase, while most existing cascade ranking in search only models the click behavior. Based on these observations, a novel cascade ranking model is designed and deployed in an operational e-commerce search application. An extensive set of experiments demonstrate the advantage of the proposed work to address multiple factors of effectiveness, efficiency and user experience in the real-world application.
\end{abstract}

%
%


\maketitle

\section{Introduction}

Many information retrieval and data mining applications such as search and recommendation need to rank a large set of data items with respect to many user requests in an online manner. There are generally two issues in this process: 1). Effectiveness as how accurate the obtained results in the final ranked list are and whether there are a sufficient number of good results; and 2). Efficiency such as whether the results are obtained in a timely manner from a user's perspective and whether the computational costs of ranking is low from a system's perspective. For large-scale ranking applications, it is important to address both issues for providing good user experience and achieving a cost-saving solution.

While most research in learning to rank for search only focuses on the effectiveness issue (e.g., ~\cite{Liu_2009, elkan2008learning, chapelle2011yahoo}), some valuable prior research of learning to efficiently rank  (e.g., ~\cite{Wang_2014, Wang_2010a}) has been proposed to address the trade-off between effectiveness and efficiency. Some previous research utilizes a feature selection process that considers both effectiveness and efficiency to speed up existing ranking models (e.g., with some time constraints of the ranking process)  ~\cite{Wang_2010a, Wang_2010b}. However, this approach applies a single ranking model with selected features to all items for a search query, which is difficult to achieve a good trade-off between effectiveness and efficiency. The cascade ranking model is proposed (e.g., ~\cite{Wang_2011, raykar2010designing, Chen_2012}) to address this problem by using a sequence of increasingly complex ranking models to progressively filter data items and refine the ranking order. The intuition is to apply ranking models with cheap features to eliminate most irrelevant data items in the beginning stages and use accurate but more expansive features to obtain accurate ranking in the later stages. Most existing research in cascade ranking is conducted in a relatively small computing environments (e.g., a few servers) with simulated user queries (e.g., from benchmark datasets), and thus do not capture multiple factors of effectiveness and efficiency in real-world applications.

This paper proposes new research work of designing and developing a cascade ranking model for a real-world large-scale e-commerce search application. To provide both good user experience and a cost-saving solution for the e-commerce search application, the new work considers multiple factors of effectiveness and efficiency as the search accuracy with respect to both the click behavior and purchase behavior, the search latency, the size of results and computational costs. In particular, a differentiable penalty function has been introduced to represent the desired properties of search latency and rank results: zero loss for small search latency (e.g., 150 milliseconds) with a large enough result size (e.g., 200) and increased loss for larger latency with smaller size. The purchase behavior and click behavior are also modeled with different weights in the overall objective function and all these factors are jointly optimized to achieve desired ranking results. An extensive set of experiments in the operational environment were conducted to demonstrate the advantage of proposed work and carefully analyze the contribution of individual factors. To our best knowledge, this is the first piece of public research work that designs and deploys a cascade ranking model for operational e-commerce search, which provides search results of millions of items (i.e., products) for hundreds of millions of users. Furthermore, we plan to make our contribution for public research in learning to rank efficiently of search (e-commerce) by sharing an appropriate version of our benchmark dataset if the paper is accepted.

The rest of the paper is organized as follows. Section 2 provides a literature survey of most related previous research work. Section 3 proposes the new cascade ranking model for e-commerce search with multiple factors of effectiveness and efficiency. Section 4 describes the experimental methodology and Section 5 presents detailed experimental results and analysis. Section 6 concludes and points out some possible further research directions.

\section{Related Work}

Machine learning algorithms such as learning to rank have been successfully used in many ranking problems with a set of available items such as Web search, feed ranking,  recommendation and product search ~\cite{chapelle2011yahoo, linden2003amazon, Liu_2009, agarwal2014activity}. Many existing learning to rank methods in search just focus on effectiveness. However, these methods may involve complicated features and models (e.g., neural network with many layers), which are often difficult to calculate in a timely manner for hundreds of millions of available items  in real-world applications.

Some work of search efficiency have also been proposed ~\cite{bayardo2007scaling, Geng_2007, lefakis2010joint, raykar2010designing, Wang_2010a} as the data size of real-world applications is getting bigger and bigger. Feature selection (e.g., ~\cite{Geng_2007}) is a common strategy to improve efficiency in learning to rank. In particular, a set of important features can be selected from all available features with respect to some criterion such as their performance in ranking to improve the ranking efficiency. The features can be selected in a query-independent manner (i.e., the same set of features for all queries ~\cite{Geng_2007}) or query-dependent manner (i.e., different set of features for different queries ~\cite{Wang_2010a, Wang_2010b}). However, the feature selection methods still calculate the same set of features for all the available items in a query and thus may not achieve a good trade-off between effectiveness and efficiency as efficient (i.e., cheap) and ineffective features can result in inaccurate results and effective and costly features can be inefficient for all items.

Cascade learning is an alternative strategy that can achieve better trade-off. It utilizes a sequence of functions in different stages and allows using different sets of features for different data items.  It can eliminate irrelevant items (e.g., for a query) in the earlier stages with simple features and models and better discern more relevant items in later stages with more complicated features and models.

Cascade learning was first applied with traditional classification and detection problems such as fast visual object detection in ~\cite{viola2001rapid, schneiderman2004feature, bourdev2005robust}. Viola et al. ~\cite{viola2001rapid} applied a series of sub-windows to sequentially reject negative examples of the whole face recognition data set and each sub-window is an Adaboost classifier with increasingly more complex features. This method allows most of the background regions of the image to be discarded quickly while consuming most computation on promising object-like regions. As this cascade approach can greatly reduce the recognition time while keeping the same accuracy, it has been introduced to various cost-sensitive scenarios ~\cite{ji2007cost, sheng2007partial}.

Cascade learning has also been applied in ranking applications for achieving high top-k rank effectiveness in an efficient manner.  The work in ~\cite{Wang_2010a, Wang_2011} uses Adaboost style framework with two functions in each stage as one function prunes the input ranked documents and the other refines the rank order. An alternative to the Adaboost cascade ranking framework is the noisy-AND probability approach ~\cite{lefakis2010joint, raykar2010designing} in which whether a certain item can go from a stage to the next depends on a simulated probability rather than a threshold. The final probability that an instance is classified as positive is the joint of the output probabilities of passing all stages. They refer to their work as `soft cascade' in which they train all the stages jointly and each stage is an independent logistic regression model. The order of the stages will not affect the effectiveness of the approach but is rather crucial when considering the efficiency. However, most existing cascade learning research for ranking is conducted in simulated environments with a small number of sampling user data/queries and a few servers. Also, current cascade learning research only considers two factors of search accuracy and total computational costs in simulated environments, while operational search applications involve other factors such as search latency and size of search results.

There are some other approaches  for improving efficiency in ranking such as caching ~\cite{Baeza_2007} and index pruning ~\cite{Carmel_2001}. The caching approach can cache results of common queries or lists of query term postings. The index pruning approach creates compact indexing structure offline (e.g., remove unimportant text terms) and allows more efficient search over the structure. These approaches are complementary with our proposed search. The research in this work focuses on ranking model and can benefit from these approaches.

\section{Algorithm Description}

Taobao is one of the largest e-commerce retail platforms with hundreds of millions of users and billions of products. Due to the huge volume of user queries and available items, it is critical to provide an  e-commerce search solution that is both effective and efficient. This section presents our solution as Cascade model in Large-Scale Operational E-commerce Search (CLOES).

Some important notations are as follows. $\mathcal{A}$ is the set of all data items in the database. When a user types a query $q$, a subset of items $\mathcal{A}_{q} = \{a_{q,i}, i=1,2,\cdots,M_{q}\} \subset \mathcal{A}$ is recalled. Let  $x_{q,i} \in \mathbb{R}^{d_x}$ denote the corresponding features of $i^{th}$ item under query $q$. Most of these query-item features are computed online and are associated with different computational costs. Besides, each query $q$ alone, can be represented as a $d_q$-dimensional vector $g(q)$. By sampling a set of $N$ instances from the online log, we get a set of training instances $\mathcal{D} = \{(x_i,q_i,y_i), i=1,2,\cdots,N\}$, where $y_i \in \mathcal{Y}=\{0,1\}$ is the label denoting if the item $x_i$ is clicked or purchased under query $q_i$ \footnote[1]{through $x_i$ and $x$ actually denote features of a certain query-item pair in the training set, we also use them as the item itself for convenience as it is easy to disambiguate from the context}. Some important query-item joint features and query-only features  can be found in table \ref{Feature}.

CLOES is treated as a typical supervised learning problem for binary classification (i.e., prediction of click or purchase). In particular, CLOES addresses multiple factors of user preferences or constraints in ranking such as search accuracy, search latency, size of search results and total CPU cost, while existing work in learning to rank only addresses one or two factors. Furthermore, CLOES improves the effectiveness of e-commerce search by modeling multiple types of user behaviors such as click and purchase, while most existing cascade ranking in general search (e.g., Web search) only models the click behavior.

\subsection{Query-Dependent Trade Off Between Effectiveness and Efficiency}

\begin{figure}
	\centering
	\includegraphics[width=0.8\linewidth]{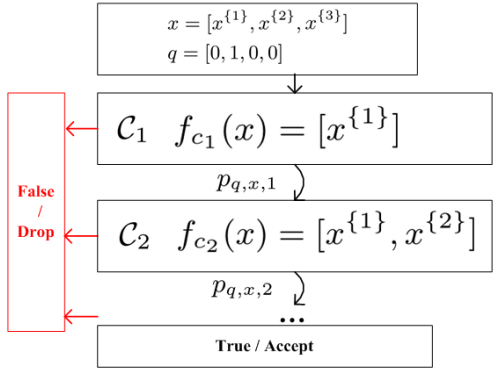}
	\caption{Cascade procedure}
	\label{fig:Cascade}
\end{figure}

This subsection describes a general architecture for constructing a cascade of classifiers which achieves both effectiveness and efficiency in the scenario mentioned above. Let $[\mathcal{C}_1,\mathcal{C}_2,\cdots,\mathcal{C}_T]$ denote a $T$ stage cascade,  where each stage $\mathcal{C}_i$ is an independent classifier using a subset $f_{c_j}(x)$ out of all features $\{x^{\{k\}},k=1,2,\cdots,d_x\}$. $f_{c_j}$ is a selecting operation in the $j^{th}$ stage. As the features used in each stage is fixed, the time/CPU cost $t_j$ of an instance in stage $j$ is also fixed. An instance is classified as positive only if all stages predict it as positive, and a negative instance can be rejected from any stage. In general, a few efficient features can be used in the former stages for quickly eliminating irrelevant items and more precise features (maybe expansive) can be used in the later stages for providing a good ranking. The cascade architecture is shown in Figure \ref{fig:Cascade}.

\begin{table*}[t]
	\centering
	\caption{Feature information}
	\begin{tabular}{c|c|c|c}
		\hline
		type& feature name & description & cost \\
		\hline
		
		\multirow{3}[1]*{Statistical}&Sales Volume & The sales volume of a certain item & 0.02 \\     \cline{2-4}
		
		&\multirow{2}[1]*{PostPay Score}& A comprehensive score after sales with respect to return rate, & 	\multirow{2}[1]*{0.09} \\
		
		& &  customer reviews and consumers ratings of an item &\\  \hline

		\multirow{6}[1]*{Predict}&\multirow{2}[1]*{Click-Through-Rate}& A prediction of click through rate on items using a logistic & 	\multirow{2}[1]*{0.13} \\
		
		&   & regression model, considering features of users, items, query and their interaction &\\ \cline{2-4}
		
		& \multirow{2}[1]*{Relevance Score} & A relevance score measuring the relevance between the query and a certain & \multirow{2}[1]*{0.74} \\
		& & item considering semantic matching features and query rewriting &  \\ \cline{2-4}
		
		& \multirow{2}[1]*{Deep \& Wide Network }& A composite score training by wide\&deep neural network ~\cite{cheng2016wide} considering  & \multirow{2}[1]*{0.84} \\
		&  & query, user action, item feature, and so on. It's accurate but rather costly &  \\ \hline
		
		& ... & ... & ... \\ \hline    \hline
		
		\multirow{2}[1]*{Query-only}& \multirow{2}[1]*{Recalled Item Count}& The range of number of recalled items. One-hot query-only feature  & - \\
		& &  that does not affect the result order but determines the size of each stage  &  \\
		\hline 		 	 	 	
	\end{tabular}
	
	\label{Feature}
\end{table*}

In this paper, a logistic sigmoid function is used for a single stage classifier

\begin{equation}
p_{q,x,j} = \sigma(w_{x,j}^T f_{c_j}(x) + w_{q,j}^T g(q))
\end{equation}

where $p_{q,x,j}$ is the probability of $x$ predicted as positive in the $j^{th}$ stage under query $q$. Different dimensions of $x$ are chosen in different stages and $q$ is used in all stages. $\sigma(z) = 1/(1+exp(-z))$ is the standard sigmoid function. Thus the cascade classifier can be described as:

\begin{eqnarray}
p(y=1|q,x) &=& p_{q,x} = \prod_{j=1}^T p_{q,x,j}
\\
p(y=0|q,x) &=& 1 - p_{q,x} = 1 - \prod_{j=1}^T p_{q,x,j}
\end{eqnarray}


Define the probability of the $i^{th}$ instance in $\mathcal{D}$ being positive as $p_i = p(y=1|q_i,x_i) = \prod_{j=1}^T p_{q_i,x_i,j}$ in all $T$ stages. Assuming that the training instances are independent, the log-likelihood function can be written as
\begin{eqnarray}
l(w) =log p(\mathcal{D}|w) = \Sigma_{i=1}^{N} y_i log p_i + (1-y_i) log(1-p_i)
\end{eqnarray}
Furthermore, the $l2$-norm is utilized to address the potential over-fitting and ill-condition issues, the objective function is changed to
\begin{eqnarray}
L_1(w) =-[\Sigma_{i=1}^{N} y_i log p_i + (1-y_i) log(1-p_i)] + \alpha \Arrowvert w \Arrowvert_2
\end{eqnarray}

\subsection{Multiple Factors of User Experience and Computational Costs}

The objective of $L_1$ focuses on ranking effectiveness. Real-world ranking applications need to address other important factors of user preferences or constraints such as search latency, size of search results and total CPU cost. This subsection presents novel research to simultaneously model these important factors.

First, the computational costs (e.g., CPU cost) are modeled with respect to the cascade principle that many irrelevant items are filtered out in the earlier stages with simple features/models and a small set of relevant documents are finely ranked in later stages with more complex features/models.
Equation (6) gives the probability that $x$ can pass the first $k$ stages.
\begin{eqnarray}
p_{q,x,pass_k} & =& \prod_{j=1}^k p_{q,x,j}   \\
\mathbb{E}[Count_{j}] & \approx& \sum_{i=1}^{N} p_{q_i,x_i,pass_j}
\end{eqnarray}

Equation (7) calculates the expected number of items can pass $j$ stages.

With equations of (6) and (7), the total expected cost can be written as:

\begin{eqnarray}
\mathcal{T}(w) = \sum_{j=0}^T \mathbb{E}[Count_{j}] \times t_{j+1}
\end{eqnarray}
Thus the objective with trade-off controller is as follow:
\begin{eqnarray}
L_2(w) = -l(w) + \alpha \Arrowvert w \Arrowvert_2 + \beta \mathcal{T}(w)
\end{eqnarray}

Additionally, similar to (7), the expected number of items that can pass the first $j$ stages under a query is:

\begin{eqnarray}
\mathbb{E}[Count_{q,j}] \approx \frac{M_q}{N_q} \times \sum_{i=1}^{N_q} p_{q_i,x_i,pass_j}
\end{eqnarray}
where $N_q$ is the number of items under query $q$ in $\mathcal{D}$ and $M_q$ is the number of recalled items in the search engine. This expected number in Equation (10) is served as the threshold for filtering out items in the corresponding stage.

The optimized result of (9) is still not desired for real-world applications as it may lead to unsatisfactory user experience. This is mainly due to two factors:  1) For a long tailed query, the cascade learning process may drop most items in the earlier stages and only a few are left in the final result; 2). For a hot query with many relevant items, it may take a long time to compute the features of millions of items and the process may result in unsatisfactory search latency. To address these two problems, this subsection presents new research of incorporating two penalties of generating too short ranked lists and too much search latency.

The first penalty term ensures that the number of search results is large enough. For example, search engine is expected to return at least $N_o$ items for a query. When we assume that the online data distribution is the same as it is in the training data , the modified objective can be written as follow:
\begin{eqnarray}
\begin{split}
&\hat{w} = arg \inf_{w} \sum_{i=1}^N \xi_i + L_2(w)
\\
s.t. &\hspace{0.5cm} Count_{q_i,T} \geq N_o - \xi_i, \hspace{0.5cm} \xi_i \geq 0
\end{split}
\end{eqnarray}

where $Count_{q_i,T}$ denotes the number of items in the final cascade stage, and the expected value can be presented by Equation (10). By eliminating $\xi_i$ the above formulation becomes:
\begin{eqnarray}
&\hat{w} = arg \inf_{w} \sum_{i=1}^N g(Count_{q_i,T}, N_o) + J(w)
\end{eqnarray}
where
\begin{eqnarray}
g(z, N_o) = max(N_o - z, 0)
\end{eqnarray}

However, (12) is similar to the hinge loss of a Support Vector Machine (SVM) model and is difficult to be directly optimized as the penalty term is not differentiable. A smooth approximation with a modified logistic regression formulation is introduced instead of $g(z, N_o)$:
\begin{eqnarray}
&g^{'}(z, N_o)=\frac{1}{\gamma}ln(1+exp(\gamma (N_o-z)))
\end{eqnarray}
Equation (14) is a differentiable function, but has a nice property of approximating the hinge loss. The three curves of SVM hinge loss $g(z)$, the logistic loss $ln(1+exp(-z))$ and the modified logistic loss $\frac{1}{\gamma}ln(1+exp(\gamma (1-z)))$ are plotted in Figure 1. The shapes of the functions are similar. It can be proofed that the gap between the modified logistic loss and the hinge loss can be eliminated with a large value of the scaling the factor $\gamma$.	
The second penalty term can be constructed in a similar way to ensure that the search latency does not exceed a desired value as ${T_l}$. Both two penalty terms can be introduced into the objective and the final loss function is:

\begin{equation}
\begin{split}
L_3(w) &= -l(w) + \alpha \Arrowvert w \Arrowvert_2 + \beta \mathcal{T}(w) \\
+ \delta \sum_{i=1}^N & g^{'}(Count_{q_i,T}, N_o) +
\epsilon \sum_{i=1}^N g^{'}(T_l, Latency_{q_i,T})
\end{split}
\end{equation}

where $T_l$ is the online latency threshold and $Latency_{q_i,T}$ is the expected latency of $q_i$ as:

\begin{eqnarray}
\mathbb{E}[Latency_{q_i,T}] = \sum_{j=1}^{T} t_j \times \mathbb{E}[Count_{q,j}]
\end{eqnarray}

In order to minimize $L_3(w)$,  the Stochastic Gradient Decent (SGD) algorithm is utilized because of its simplicity, speed, and stability. The parameters are first initialized to be random values around zero and then moved to the direction according to the gradient. $\beta$ is tuned to get the best performance under the limited CPU cost while $\delta$ and $\epsilon$ are tuned to ensure user experience.

%

\begin{figure}[t]
	\centering
	\includegraphics[width=7cm,height=4.5cm]{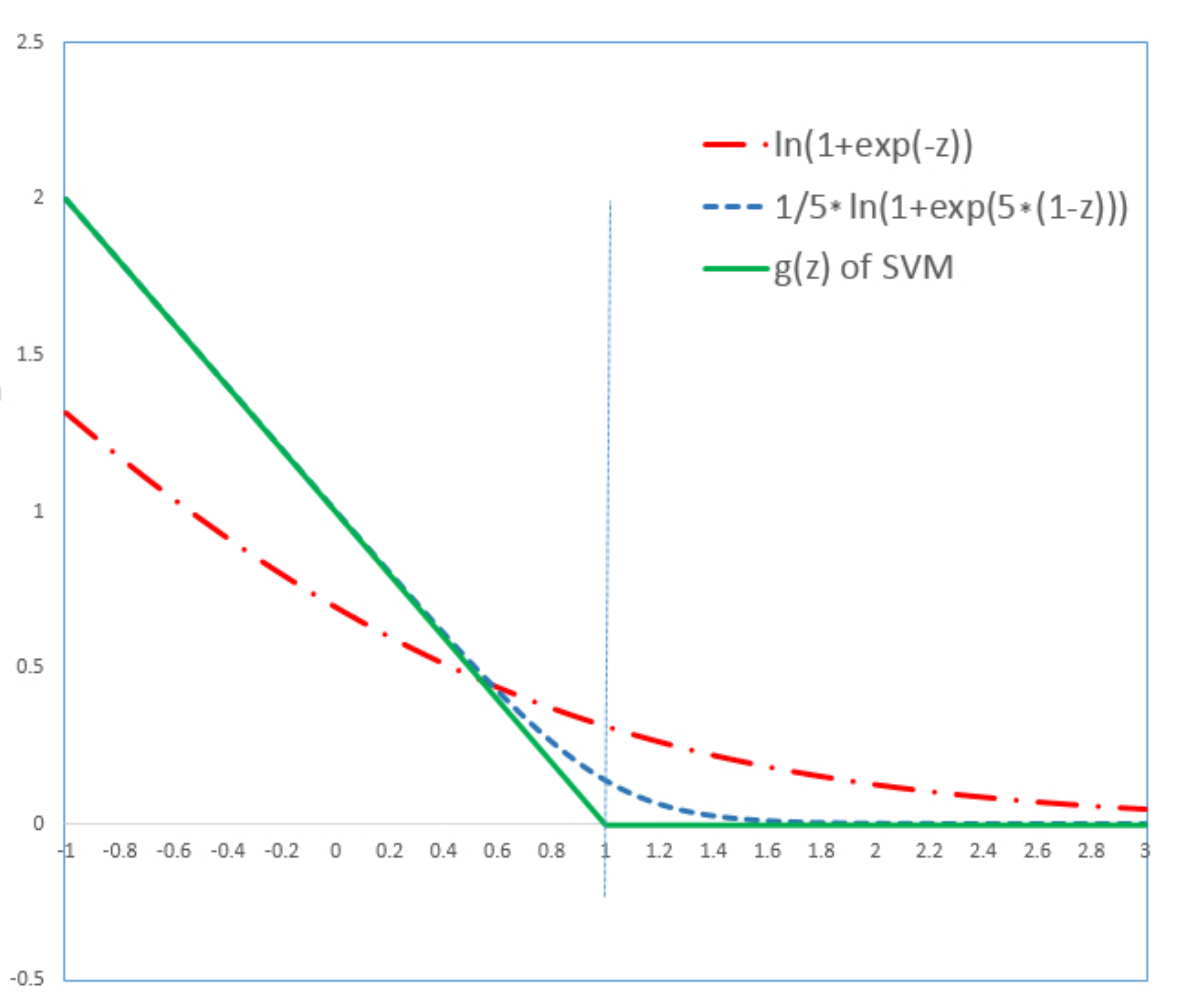}
	\caption{Loss function of SVM, logistic regression and the modified logistic regression}
\end{figure}

\subsection{Importance Factors of E-commerce Search}
\label{sec::behav}
Most existing cascade ranking in search only models the click behavior, while the effectiveness of e-commerce search involves multiple types of user behaviors such as click and purchase. Furthermore, the benefit of different purchase transactions (e.g. the contribution of Gross Merchandise Volume as GMV) to an e-commerce platform can be different based on different prices of purchased items (e.g. a refrigerator or an ice cream) and thus different purchase transactions should also be treated differently. As positive examples of clicked items are much more common than positive examples of purchased items, the optimization goal favors more on click through rate(CTR) than GMV if all data are treated equal. So importance weights are introduced on different types of user behaviors and items with different prices, specifying the relative importance. In particular, the purchased items are weighted to be $\varepsilon$ times more important than clicked examples and each behavior of an item is weighted by the product of a constant  $\mu$ and $log(price)$ for reflecting the item price. This means a purchased instance of an item with high price can be much more important than a clicked example of a cheap item. More specifically, the log-likelihood function of (4) changes to:

\begin{equation}
\begin{split}
l(w) &= \Sigma_{i=1}^{N} [y_i log p_i + (1-y_i) log(1-p_i)] wgt_i \\
wgt_i &= \left\{
\begin{aligned}
& \varepsilon \times \mu \times log(price) && if \ x_i \ is \ purchased \\
& \mu \times log(price) && else \ if \ x_i \ is \ clicked \\
& 1  && no \ behavior \ on \  x_i
\end{aligned}
\right.
\end{split}
\end{equation}

where $y_i$ denotes the label of a instance. It's set to 1 when $x_i$ is clicked or purchased and it's set to 0 otherwise. The optimization process utilizes these importance weights in the gradient descent procedure.

\section{Experimental Methodology}
\label{sec::Exp_Meth}
%
%

This section presents the experimental methodology. First, it describes the off-line evaluation setting and the online operational environment. Second, it provides a summary of other algorithms to be studied with the proposed research.

\subsection{Experimental Environment}

\begin{itemize}
	\item{Off-line Benchmark Dataset}
\end{itemize}

The off-line benchmark dataset is drawn from the search log of Taobao in a day of late October, 2016. It's randomly sampled from the online e-commerce search log where each data instance contains the user, the query, the item, the corresponding features and the match-count that stands for the number of recalled items for a specific query. In total, 2 million instances are collected with dozens of features. The ratio between positive and negative instances per query is about 1 to 10. The positive instances denote the user behaviors of click or purchase and negative instances indicate that the users did nothing on the corresponding instances. In experiments on the off-line data, different behaviors are not distinguished.

The detailed information of some important features and associated computational costs are listed in table \ref{Feature} (Due to the page limitation, not all features are listed.), where the first 5 features are query-item features and the last one-hot feature is the query only feature. Query-item features can be further divided into 2 types: statistical features and predictive features. Statistical features describe the basic item statistics such as the sales volume and sales service score. Predictive features are scores calculated by different machine learning models for various targets, e.g. CTR prediction by logistic regression, preference by collaborative filtering and session contribution by deep deterministic policy gradient. It can be seen that some relatively cheap features can be easily calculated such as sales volume but their performance in rank may be not high, while some more complicated features such as Deep \& Wide Network Score can be more accurate but more expensive to compute. The query only feature with the count of recalled items is used to control the magnitude of the prediction probability (thus to control the result number and cost per query) but does not affect the rank order. The full data set is randomly divided into 5 parts for cross validation.

\begin{itemize}
	\item{Online Environment}
\end{itemize}

The search engine of Taobao is a rather big and complex system with billions of items and hundreds of millions of user queries everyday. Usually there are about 40 thousand queries per second supported by two clusters (for potential accidents) with hundreds of computers in a distributed search environment, and the query volume could be tripled during peak periods of some promotional campaign such as the Singles' Day shopping festival. Therefore, system efficiency has always been an extremely important problem. Search experience mainly means that it is highly desired to show a sufficient number of good products to consumers within a limited latency, and the computational cost of the system can be defined as the CPU cost. The desired property of the operational e-commerce search system is set according to our empirical experience: the search system should respond in no more than 130 ms, the number of returned products should be at least 200, and the average CPU utilization rate in computer clusters should be  below 70\%.

A standard A/B test is conducted online, where about 5\% of the random users are selected for testing. Some system information of each computer in the two clusters is listed in the following table \ref{table:system}.

\begin{table}[h!]
	\centering
	\caption{System information}
	\begin{tabular}{ll}
		\hline
		hardware&configuration\\
		\hline
		\multirow{2}[1]*{CPU}&  2x 16-core Intel(R) Xeon(R)  \\
		&CPU E5-2682 v4 @ 2.50GHz\\
		RAM&  256 GB\\
		Is hyperthreading&  Yes\\
		Networking&  10 Gbps\\
		OS&  ALIOS7 Linux 3.10.0 x86\_64   \\
		\hline
	\end{tabular}
	
	\label{table:system}
\end{table}

\subsection{Algorithms for Comparison}
The following set of algorithms are studied in the paper for providing a comprehensive set of results.

\begin{itemize}
	\item{\textbf{Single stage classifier} is a common choice where all the features are used only once in a single stage. In our experiments, it's a logistic regression model. If it contains all the features of other algorithms in comparison, the single big classifier can be very accurate but quite costly and slow as all the features have to be calculated for all items. It is the baseline algorithm of the experiments and at most time it is too expensive, as majority of negative items are so obvious that some complicated features are not necessary. If only cheapest features (e.g., sales volume) are selected, the classifier can be much faster but inaccurate.}

	\item{\textbf{2-stage approach} is a heuristic and widely used algorithm in real large-scale search or recommender system (e.g., ~\cite{covington2016deep}). It is a simple and efficient approach that has been used in the Taobao search system in the past several years. In the first stage, all recalled items are filtered by some cheap features manually selected by human supervision. In our system (and in the experiments), regularized sales volume is the only chosen feature in this stage and the number of items that can pass through the first stage is set as a constant of 6000. In the second stage, the remained items are ranked with the rest of features. There are more than 40 features of each query-item pair, and this number is still increasing. However, this simple approach does not provide a formal guidance of how to select features in the two stages and how to set the number to determine how many items can pass into the second stage for achieving a good trade-off between effectiveness and efficiency. Though it is possible to apply more than 2 stages in this approach, we find it difficult to manually search for a good combination of size thresholds of the stages and feature assignment in each stage for obtaining desired performance. }
	

	\item{\textbf{CLOES} is the algorithm proposed in this paper. The recalled items are ranked in a cascade way. It considers multiple factors of consumer experience and system effectiveness and efficiency in ranking. Taobao search system now applies the CLOES of 3 stages. To our best knowledge, this is the first piece of public research work that designs and deploys a cascade ranking model for operational e-commerce search. Queries are distinguished in our work and the off-line evaluation cost is quite consistent with the online cost.  }

\end{itemize}

\section{Experiment and Analysis}
\subsection{Off-line Comparison}
The first set of experiments are conducted off-line to compare CLOES with some alternative algorithms on the benchmark dataset described in Section ~\ref{sec::Exp_Meth}. The purpose is to investigate the rank effectiveness as Area Under the Curve (AUC) and rank efficiency as CPU cost in the offline manner. The CPU cost of the single stage classifier (logistic regression) with all features is set to 1 as the baseline cost and the cost of the other approaches are presented as the ratio of the baseline cost. The results are shown in table \ref{methods compared with off-line benchmark}.

It can be seen from the table that the single stage classifier with all features get the highest accuracy, but at the same time it has the largest computational cost as all the features have to be calculated for all items. On the other side, the single stage classifier with cheapest features has the lowest AUC (about 15\% lower) with the least computational cost (about 6\% of single stage classifier with all features). Both two variants do not represent a good trade-off solution as the former one is hard to be applied online due to the expensive CPU cost and the later one is not desired to be used due to its poor accuracy.

The 2 stage approach has a lower accuracy than the single stage classifier with all features, but is considerably more efficient. The performance of this approach depends on the size of result set from the first stage (i.e., how many items can pass the first stage), which is manually tuned from 2000 to 8000. A bigger result set is associated with better accuracy but more computational cost. More specifically, it is set to 6000 in the Taobao search engine for a good trade-off result. The soft-cascade algorithm is also evaluated. It performs in a similar manner to the 2 stage approach, with a slightly better average accuracy and a slightly lower average CPU cost. However, the online efficiency of soft cascade is less predictable than the 2 stage approach with a fixed size of the result set from the first stage and thus may result in bad user experience (i.e., too much search latency) due to a too big results set from the first stage.

Experimental results of the proposed CLOES with different values of the hyper-parameter $\beta$ can be seen from table \ref{methods compared with off-line benchmark}. The $\beta$ parameter is introduced to balance the accuracy and the CPU cost in search while a larger value of $\beta$ puts more weight on the efficiency issue. The ideas of CLOES and the 2 stage approach are similar. The 2-stage approach filters out many irrelevant items in the first stage and finely tunes the rank of other items in a more expensive second stage while CLOES ranks the items with increasingly more complex features in multiple stages. However, CLOES learns to reach a better trade-off with a more formal approach while the 2 stage approach is heuristically set by human intuition. The results in table \ref{methods compared with off-line benchmark} indicate that CLOES can get a better accuracy (AUC, from 0.76 to 0.81) with the same CPU cost as $\beta$ equals to 1 and get a higher accuracy with much less CPU cost (from 0.30 to 0.18) as $\beta$ equals to 10.

In summary, this set of experiments show that CLOES can achieve a nice trade-off with good accuracy and low CPU cost on the off-line benchmark dataset. However, offline evaluation does not consider the full spectrum of feedback from real-world users (e.g., the effects from large search latency and small result set) in an operational environment.

\subsection{Online Evaluation in Operational Environment}

\begin{table}[t]
	\centering
	\caption{Methods Compare}
	\begin{tabular}{llll}
		\hline
		& Train set & Test set & Cost \\
		& AUC  & AUC & COST \\
		\hline
		Single stage &\multirow{2}[1]*{0.88} & \multirow{2}[1]*{0.87}   & \multirow{2}[1]*{1} \\
		(all features) &   &   \\
		
		Single stage & \multirow{2}[1]*{0.73}   & \multirow{2}[1]*{0.72} & \multirow{2}[1]*{0.06} \\
		(simple features) &    &  &  \\
		
		2-stage approach & 0.78   & 0.76 & 0.30 \\
		
		CLOES($\beta=1$) & 0.81 &   0.80 & 0.29 \\
		
		CLOES($\beta=10$) & 0.80 &   0.77 & 0.18 \\
		\hline
	\end{tabular}
	\label{methods compared with off-line benchmark}
\end{table}

This subsection and the following subsections present the results of online evaluation in the real-world operational environment of Taobao e-commerce search with a standard A/B testing configuration. The parameter $\delta$ and $\epsilon$ were tuned empirically to 1 and 0.05. In particular, the results in this subsection and the results in subsection \ref{sec::cust} discuss results in usual business days, while the results in subsection ~\ref{sec::d11} are during a major promotion period (i.e., Singles' Day shopping festival).

Results of the 2 stage approach is used as the baseline as this was the operational solution in Taobao search before the proposed research work. Different variants of the proposed CLOES with respect to different types of user behaviors (i.e., click and purchase) are compared.  Subsection \ref{sec::behav} introduces the $\epsilon$ parameter to represent the relative weight of a purchase behavior against a click behavior and the parameter $\mu$ to represent the importance of an item with a higher price. These variants of CLOES utilize different values of the two parameters for thoroughly investigating the performance of CLOES in the operational e-commerce search environment. The results are shown in table \ref{resample}.

\begin{figure}[h]
	\centering
	\includegraphics[width=0.8\linewidth]{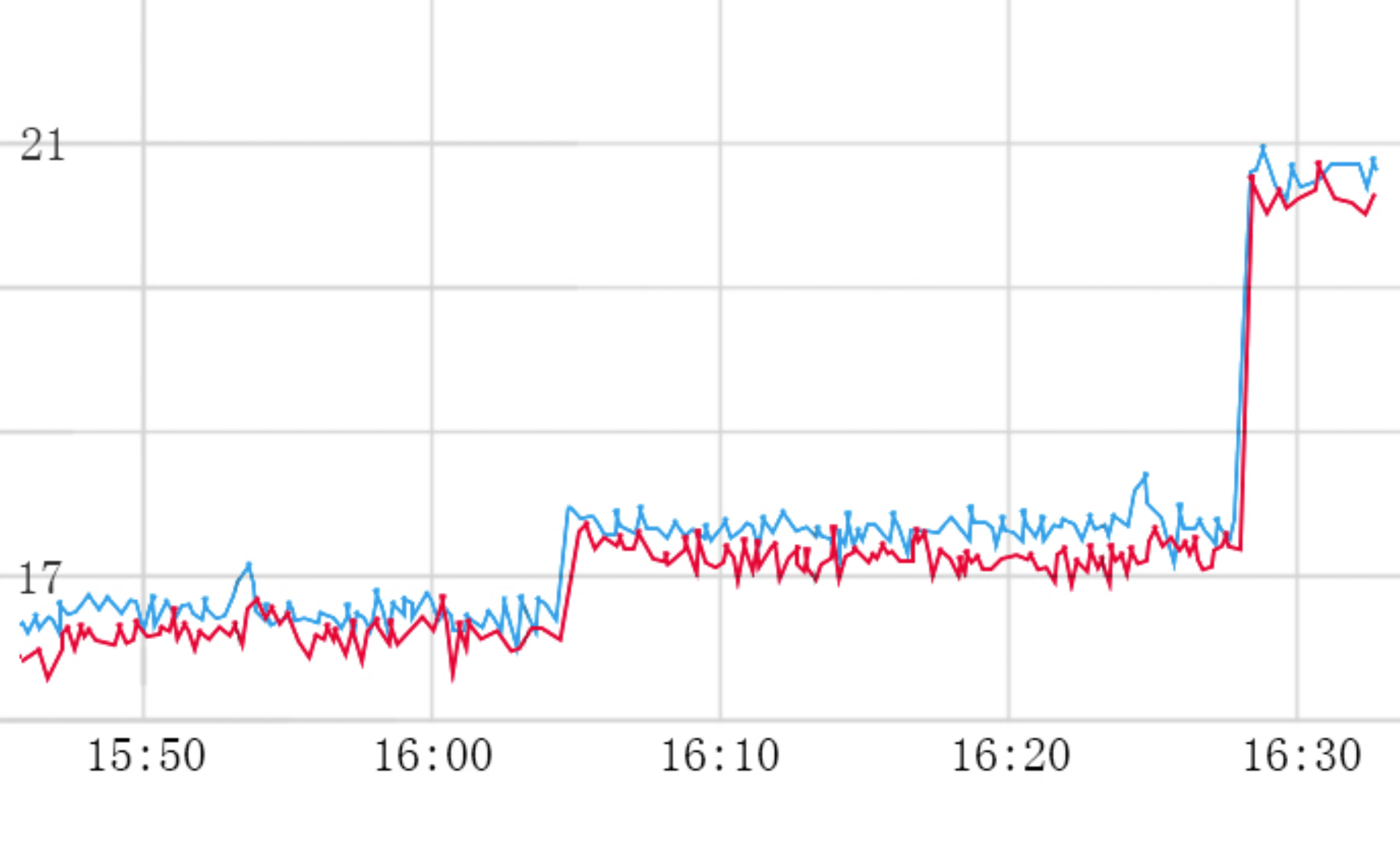}
	\caption{Change of rank search latency when uninstall CLOES}
	\label{fig:rank-search-latency}
\end{figure}

It can be observed from the table that CPU cost of all variants is 20 \% lower than the 2 stage approach as the $\beta$ value is set to 5 for favoring an efficient solution. The first line of the table is the variant of CLOES without distinguishing the click behavior and the purchase behavior by setting $\epsilon$ to 1. However, more attention should be paid to the purchase behavior than the click behavior for an e-commerce platform. Therefore, the variants of line 2 to line 5 use a higher value of $\epsilon$ as 10 as each purchased item is viewed 10 times more important than a clicked item. It can be seen that both the metrics of number of orders and Gross Merchandise Volume (GMV) improve for these variants than the first variant while the Click Through Rate (CTR) drops a bit. This is consistent with our expectation as the new variants put more emphasis on the transaction data (i.e., purchase and price) than the first variant.  From line 2 to line 5, different values of the parameter $\mu$ are applied to study the effect of item price. Items with a higher price tend to rank higher in the final ranked list as the value of $\mu$ increases. The CTR metric and the number of orders metric decrease a bit from line 2 to line 5 as some users may not be interested in products with higher prices, but the GMV metric increases first and then drops, while each transaction can generate more value but may be of less interest to users.  It seems that the setting in line 4 (i.e. $\epsilon$ as 10 and $\mu$ as 3) provides the highest GMV value. However, different settings may be more desired for different contexts (e.g., CTR may be important if users' interaction is viewed more important).

\begin{figure*}[htbp]
	\centering
	\includegraphics[width=18cm,height=9cm]{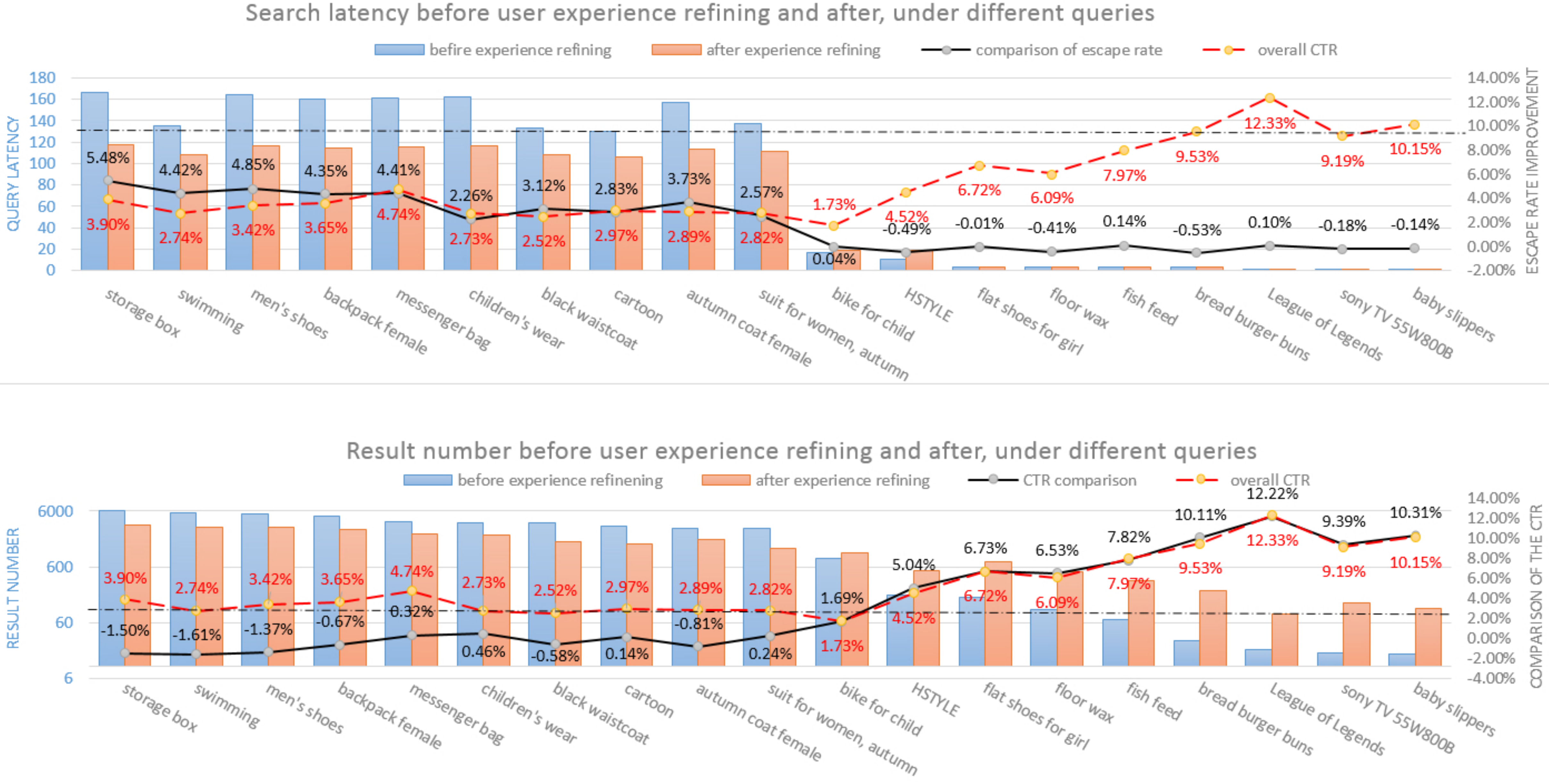}
	\caption{Performance of CLOES on different queries}
	\label{fig:user-experience}
\end{figure*}

\begin{table}[t]
	\centering
	\caption{Importance Weight(\%)}
	\begin{tabular}
		{l  |lllll}
		\hline
		Importance & \multirow{2}[1]*{CTR} & Number  & \multirow{2}[1]*{GMV} & Unit  & \multirow{2}[1]*{Cost} \\
		weight & & of order &  &  price &  \\
		\hline
		$\varepsilon=1, \ \mu=1$&+1.58& $-$1.35 & $-$1.76 & $-$0.42 & $-$20 \\
		$\varepsilon=10, \mu=1$ &+0.25& +1.89 & $-$0.64 & $-$2.49 &  $-$20 \\
		$\varepsilon=10, \mu=2$ &+0.17& +1.65 & +0.24 & $-$1.39 &  $-$20 \\
		$\varepsilon=10, \mu=3$ &+0.12& +0.36 & +1.32 & +0.95 &  $-$20 \\
		$\varepsilon=10, \mu=4$ &$-$0.13& $-$0.25 & $-$0.92 & +1.65 &  $-$20 \\
		\hline
	\end{tabular}
	
	\label{resample}
\end{table}

Figure \ref{fig:rank-search-latency} provides the search latency results of the online system when we turn off CLOES for evaluation. The figure shows two curves that indicate the average latency of two independent computer clusters. In particular, the CLOES was shutdown for switching to the original approach (2 step) in two steps.  First,  a small portion of the CLOES search traffic is switched to the 2 step approach for gray testing, and after about 20 minutes CLOES is fully turned off. It can be seen from the figure that there are two obvious rising of the latency curves with respect to the two steps, and the latency is increased from about 17 ms to 21 ms. Both this figure for online latency and the data in table \ref{resample} for online computation suggest that CLOES can save about 20\% of the CPU cost.

\subsection{Users' Experience in Operational Environment}
\label{sec::cust}

The subsection provides more experimental results and thorough analysis for studying users' experience in the large-scale operational e-commerce search application. It is important to ensure that users can get a sufficient number of good products in a limited time as represented by Equation (15). Figure \ref{fig:user-experience} shows the comparison results of user experience for 19 representative queries with and without modeling of user experience as search latency and size of search results. Hot queries (i.e., queries with more returned results and larger search latency) are listed in the left and long-tail queries (i.e., queries with fewer returned results and smaller search latency) are listed in the right. In each bar pair, the left one is from the cascade method without modeling user experience and the right one is from the complete CLOES method of modeling user experience.

The top sub-figure presents the results of the average query latency and the variation of escape rate, which is the percentage of users who immediately (i.e., less than 1 second) leave a search page. Generally speaking, the more time the search system responds, the more likely a user escapes. For hot queries, as there are many items to compute, the latency can be quite high (e.g., average 170ms for `storage box') without modeling user experience, and thus more users may give up the search operation under these queries. In contrast, it can be seen from the sub-figure that the latency of all listed queries decrease below 130ms (the dash line) with modeling user experience and the escape rate decreases accordingly. For example, the latency decreases noticeably under query `swimming' (i.e. from 170ms to 108ms) and escape rate can decrease about 5\%. On the other side, the escape rate of long-tail queries does not change much with modeling user experience as their original search latency is already low, which suggests that users are more sensitive to the latency difference when the latency is higher.

The bottom sub-figure shows the result count of each query and the variation of the query CTR without escaped queries and overall CTR with all queries. For example, under query `swimming', the result number decreases from 5700 to 3100, CTR without escaped queries decreases about 1.5\% and CTR with all queries increases about 2.5\% ; under `floor wax', the result number increases about 8 times and both the CTR metrics increase about 6.7\%. From the sub-figure, three important observations can be made: First, the result number of hot queries decreases while the number of long tailed queries are substantially increased to about 200 (the dash horizontal line) as modeled from Equation (10); Second, the query CTR without escaped queries increases a lot for long tailed queries and is about the same or a bit lower for hot queries; Third, in most cases the overall CTR improves or is stable. The observations are as expected as users usually only browse the top part of ranked items and rarely view in the bottom. So the CTR (i.e., without escaped queries ) improvement for long-tailed queries are more substantial as a large portion of good-quality results are introduced by modeling user experience while the loss of good-quality results for hot queries is minor with modeling user experience. Furthermore, modeling user experience reduces search latency for hot queries as discussed previously, and thus improves the overall CTR even for hot queries.

We also compared users' experience of CLOES with the former 2-stage approach. Table \ref{resample} illustrates that the overall query latency of CLOES decreases about 20\% and CTR can even increase 1.58\%. In the 2-stage approach, result number is manually set as a constant for all queries which is usually bigger than in CLOES in the final stage. But CTR in CLOES is higher even with fewer results because of its accuracy. Escape rate of CLOES is also lower than in the 2-stage approach due to its low latency.

\subsection{Singles' Shopping Festival}
\label{sec::d11}

Singles' Day shopping festival is an important day of Taobao and is one of the biggest online shopping festivals around the world. In 2016, the Singles' Day festival hit  close to \$18 billion dollars of sales in a single day (the contribution of search is more than 10\%) and attracted hundreds of millions of users from more than 200 different countries \footnote[1]{\url{https://techcrunch.com/2016/11/11/alibaba-singles-day-2016/}}. The peak level of trade rate reached 0.17 million per second in the first half hour. The e-commerce search system played an important role in this big event.

In Singles' Day, the search traffic of the e-commerce search engine increases about three times than in a common day. On one hand, it is desired to further improve the search accuracy in this important day by adding more real-time features, which are often effective in a dynamic period as Singles' Day but rather expensive. On the other hand, the engine is not able to support all the features under the tripled QPS. In previous years, system degradation was necessary and the degradation was often executed in two simple and naive ways: first, some costly features were dropped such as features using NN (neural network) models or large GBDT models ~\cite{friedman2001greedy}; Second, the size of candidate items in the two stage approach was decreased to almost half of a common day. Unfortunately, both two ways are rather harmful to the search accuracy.

\begin{figure}[h]
	\centering
	\includegraphics[width=0.8\linewidth]{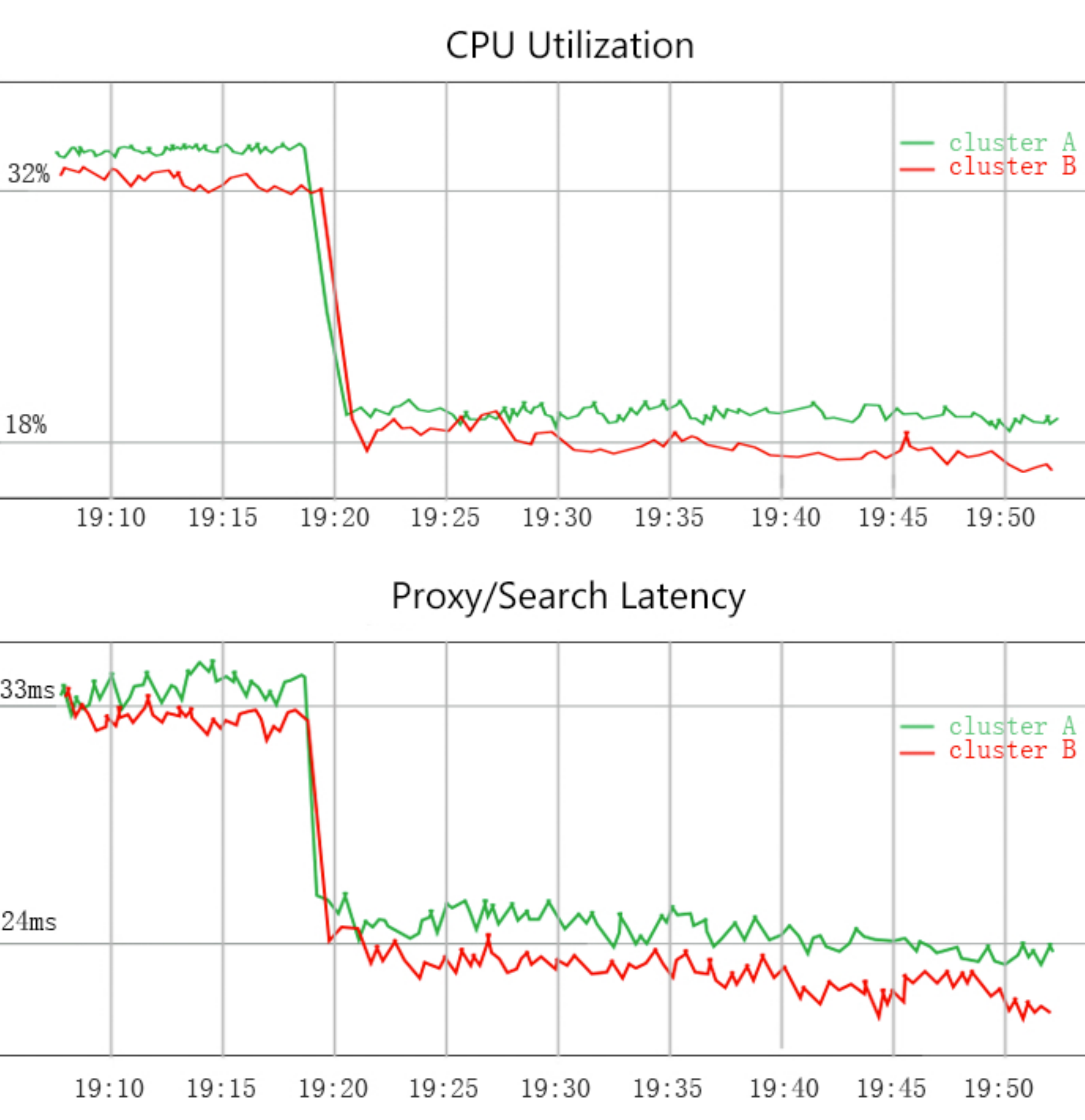}
	\caption{CPU utilization and search latency }
	\label{fig:double11}
\end{figure}

In the Singles' Day event of 2016, the search system of Taobao did not suffer from system degradation by applying the proposed CLOES approach. In particular, the trade-off between search effectiveness and CPU cost is carefully achieved by tuning the $\beta$ parameter in CLOES. To find an appropriate $\beta$, experiment was conducted a few days before November 11th, 2016, and finally we set $\beta$ as 10. The CPU utilization is expected to be equal or less than 70\% even at the evening peak periods of Singles' Day for system stability. Accordingly, it is considered safe only when the utilization is decreased to about 20\% in the experiment of a common day. Figure \ref{fig:double11} reports the change of CPU utilization and search latency during the experiments. Two sub-figures draw the CPU utilization and the average search latency respectively with each curve standing for a computer cluster. It is clear that the CPU utilization was above 32\% in normal situation without CLOES. When CLOES is applied, the CPU utilization dropped to about 18\% (45\% was saved) rapidly and the average search latency was decreased about 30\% (from 33ms to 23ms). With the same setting, the system performance (GMV) was almost the same or a little higher (0\% to 1\%).

In the very day (Nov 11th, 2016) of Singles' Day 2016, CLOES was fully applied in the Taobao search system and the CPU utilization peaked to 75\% in the evening. Benefiting from CLOES, no feature was dropped out and some real-time features were added. New real-time features achieved a 10-20\% CTR improvement in the very day and the search system performed much better than previous years in terms of search accuracy and stability.

\section{Conclusion}
This paper presents novel research of designing and deploying cascade ranking in the Taobao e-commerce search application. As far as we know. it is the first piece of public research and thorough study of designing and deploying a cascade model in a large-scale operational e-commerce search engine, which deals with hundreds of millions of user queries per day with hundreds of servers. The proposed approach models multiple factors of user experience and computational cost and addresses multiple types of user behavior in e-commerce search. An extensive set of experiments and detailed discussion are provided to show that the proposed solution provides a good trade-off between search effectiveness and search efficiency within operational environments in regular e-commerce environment and the intensive Singles' Day shopping festival .

There are several possible future research directions. First, each classifier of the current cascade is a simple linear model while more complex models may work better. Second, the current approach follows a standard setting of supervised learning, which may be extended to the reinforcement learning setting for modeling both the immediate rewards (buying the products right now) and delayed rewards (e.g., visiting or buying in the future).

\section{ACKNOWLEDGMENTS}
We thank colleagues of our teams $ - $ Yinghui Xu, Xi Chen, Heng Li, Qing Da and Yabo Ni for useful discussions and supports of this work. We thank our cooperative teams  $ - $ search engineering team and iDST . We also thank authors of prior works on balancing the ranking effectiveness and efficiency for the inspiration they gave us. We finally thank the anonymous reviewers for their valuable feedback.

\bibliographystyle{abbrv}
\bibliography{cascade_learning} 

\end{document}